%% file: root.tex
\documentclass[letterpaper, 10 pt, conference]{ieeeconf}

\IEEEoverridecommandlockouts                              % This command is only needed if 
\usepackage{xcolor}
\usepackage{graphicx}
\usepackage{xspace}
\usepackage{listings}

\usepackage{url}
\input{macros.tex}

\usepackage{amsmath}
\usepackage{amsfonts}
\usepackage{todonotes}

\usepackage{booktabs}
\usepackage{float}
\usepackage{tikz}
\title{\LARGE \bf Can LLMs plan paths with extra hints from solvers?}

\author{Erik Wu and Sayan Mitra% <-this % stops a space
\thanks{*This work was supported by a research award from the C3E Cybersecurity Challenge Problem program}% <-this % stops a space \textcolor{blue}{Mention your DAAD fellowship to visit UIUC.}
\thanks{E. Wu is with the Department of Computer Science at the Karlsruhe Institute for Technology. email: erik.wu@student.kit.edu}
\thanks{S. Mitra is with the Department of Electrical and Computer Engineering at the University of Illinois at Urbana-Champaign. email: mitras@illinois.edu}%
}

\renewcommand{\baselinestretch}{.95}

\begin{document}

\maketitle
\thispagestyle{empty}
\pagestyle{empty}

%%%%%%%%%%%%%%%%%%%%%%%%%%%%%%%%%%%%%%%%%%%%%%%%%%%%%%%%%%%%%%%%%%%%%%%%%%%%%%%%

\input{00_Abstract}

\input{01_Introduction}

% \input{02_Background}
\input{03_Method}

% \input{04_Reachability} 
\input{06_Results}
\input{07_Conclusions}
% \input{appendices}

%%%%%%%%%%%%%%%%%%%%%%%%%%%%%%%%%%%%%%%%%%%%%%%%%%%%%%%%%%%%%%%%%%%%%%%%%%%%%%%%

% ref: \cite{IEEEexample:articleetal}, \cite{IEEEexample:book}, ...

%\newpage

\bibliographystyle{IEEEtran}
\bibliography{IEEEabrv,IEEEexample,kt-bib,sayan1}
%\printbibliography
\newpage

\end{document}

%% file: macros.tex
\newcommand{\init}{\mathit{I}}
\newcommand{\goal}{\mathit{G}}
\newcommand{\obs}[1]{\mathit{O}_{#1}}

\newcommand{\W}{\mathit{W}}

\lstdefinelanguage{pseudocode}{
    basicstyle=\notsotiny,
	keywordstyle=\bf \notsotiny,
	mathescape=true,
	tabsize=20,
	xleftmargin=4.0ex,
	basicstyle=\normalsize,
	sensitive=false,
	columns=fullflexible,
	keepspaces=false,
	basewidth=0.05em,
	moredelim=[il][\rm]{//},
	moredelim=[is][\sf \figuresize]{!}{!},
	moredelim=[is][\bf \figuresize]{*}{*},
	keywords={automaton, algorithm, and, 
		break,
		choose,const,continue, components,
		discrete, do,
		eff, external,else, elseif, evolve, end, each, exit,
		fi,for, forward, from, find, 
		hidden,
		in,input,internal,if,invariant, initially, imports,
		let,
		mode,
		or, output, operators, od, of,
		pre,
		return,
		such,satisfies, stop, signature, simulation, sample,
		trajectories,trajdef, transitions, that,then, type, types, to, tasks,
		variables, vocabulary, 
		when,where, with,while},
	emph={set, seq, tuple, map, array, enumeration},   
	literate=
	{(}{{$($}}1
	{)}{{$)$}}1
	{\\in}{{$\in\ $}}1
	{\\preceq}{{$\preceq\ $}}1
	{\\subset}{{$\subset\ $}}1
	{\\subseteq}{{$\subseteq\ $}}1
	{\\supset}{{$\supset\ $}}1
	{\\supseteq}{{$\supseteq\ $}}1
	{\\forall}{{$\forall$}}1
	{\\le}{{$\le\ $}}1
	{\\ge}{{$\ge\ $}}1
	{\\gets}{{$\gets\ $}}1
	{\\cup}{{$\cup\ $}}1
	{\\cap}{{$\cap\ $}}1
	{\\langle}{{$\langle$}}1
	{\\rangle}{{$\rangle$}}1
	{\\exists}{{$\exists\ $}}1
	{\\bot}{{$\bot$}}1
	{\\rip}{{$\rip$}}1
	{\\emptyset}{{$\emptyset$}}1
	{\\notin}{{$\notin\ $}}1
	{\\not\\exists}{{$\not\exists\ $}}1
	{\\ne}{{$\ne\ $}}1
	{\\to}{{$\to\ $}}1
	{\\implies}{{$\implies\ $}}1
	{<}{{$<\ $}}1
	{>}{{$>\ $}}1
	{=}{{$=\ $}}1
	{~}{{$\neg\ $}}1
	{|}{{$\mid$}}1
	{'}{{$^\prime$}}1
	{\\A}{{$\forall\ $}}1
	{\\E}{{$\exists\ $}}1
	{\\/}{{$\vee\,$}}1
	{\\vee}{{$\vee\,$}}1
	{/\\}{{$\wedge\,$}}1
	{\\wedge}{{$\wedge\,$}}1
	{=>}{{$\Rightarrow\ $}}1
	{->}{{$\rightarrow\ $}}1
	{<=}{{$\Leftarrow\ $}}1
	{<-}{{$\leftarrow\ $}}1
	{~=}{{$\neq\ $}}1
	{\\U}{{$\cup\ $}}1
	{\\I}{{$\cap\ $}}1
	{|-}{{$\vdash\ $}}1
	{-|}{{$\dashv\ $}}1
	{<<}{{$\ll\ $}}2
	{>>}{{$\gg\ $}}2
	{||}{{$\|$}}1
	{[}{{$[$}}1
	{]}{{$\,]$}}1
	{[[}{{$\langle$}}1
	{]]]}{{$]\rangle$}}1
	{]]}{{$\rangle$}}1
	{<=>}{{$\Leftrightarrow\ $}}2
	{<->}{{$\leftrightarrow\ $}}2
	{(+)}{{$\oplus\ $}}1
	{(-)}{{$\ominus\ $}}1
	{_i}{{$_{i}$}}1
	{_j}{{$_{j}$}}1
	{_{i,j}}{{$_{i,j}$}}3
	{_{j,i}}{{$_{j,i}$}}3
	{_0}{{$_0$}}1
	{_1}{{$_1$}}1
	{_2}{{$_2$}}1
	{_n}{{$_n$}}1
	{_p}{{$_p$}}1
	{_k}{{$_n$}}1
	{-}{{$\ms{-}$}}1
	{@}{{}}0
	{\\delta}{{$\delta$}}1
	{\\R}{{$\R$}}1
	{\\Rplus}{{$\Rplus$}}1
	{\\N}{{$\N$}}1
	{\\times}{{$\times\ $}}1
	{\\tau}{{$\tau$}}1
	{\\alpha}{{$\alpha$}}1
	{\\beta}{{$\beta$}}1
	{\\gamma}{{$\gamma$}}1
	{\\ell}{{$\ell\ $}}1
	{\\TT}{{\hspace{1.5em}}}3        
}

\lstdefinelanguage{pseudocodeNums}[]{pseudocode}
{
	numbers=left,
	numberstyle=\tiny,
	stepnumber=2,
	numbersep=4pt
}

\newcommand{\A}{\mathcal{A}}

%% file: 00_Abstract.tex
\begin{abstract}
Large Language Models (LLMs) have shown remarkable capabilities in natural language processing, mathematical problem solving, and tasks related to program synthesis. However, their effectiveness  in long-term planning and higher-order reasoning has been noted to be limited and fragile. This paper explores an approach for enhancing LLM performance in solving a classical robotic planning task by integrating  solver-generated feedback. %\textcolor{orange}{Unlike many existing methods, which primarily use LLMs to assist traditional planners, our approach focuses on improving the LLMs' own ability to solve the planning tasks independently.} 
We explore four different strategies for providing feedback, including visual feedback,  we utilize fine-tuning, and we evaluate the performance of three different LLMs across a 10 standard and 100 more randomly generated planning problems. Our results suggest that the solver-generated feedback improves the LLM's ability to solve the moderately difficult  problems, but the harder problems still remain out of reach. The study provides detailed %\textcolor{orange}{
analysis of the effects of the different hinting strategies and the different planning tendencies of the evaluated LLMs.
%}

% This paper impirically investigates the effects of different hints ...

% \begin{itemize}
%     \item LLM Hype, applied to almost everything / hammer to all problems.  LLMs cannot plan paths on their own.
%     \item How man much can we help it with solver aided hints and giving feedback
%     \item By giving LLMs feedback, the LLM can reflect on its previous answer and improve upon it. And thus utilizing the reactive nature of LLMs (compared to regular deterministic computer programs)
%     \item We demonstrate that the LLM is significantly better in solving problems using this approach. However, LLMs still fail to plan spatially more challenging problems
% \end{itemize}
\end{abstract}

%% file: 01_Introduction.tex
\section{INTRODUCTION}
\label{sec:introduction}
Planning is one of the  classical robotics problems  that requires both step-by-step decision making and world knowledge~\cite{Lavalle-planning06}. 
Large Language Models (LLMs) have shown remarkable capabilities in natural language processing, mathematical problem solving, and tasks related to program synthesis, using only implicit knowledge about the world in the form of large volumes of data.
%
%There is a tantalizing interest in using LLMs, for planning  (see, for example~\cite{AGISparks23,wei2022emergentabilitieslargelanguage,LLM-Planner-23} and the references therein).
%
%
LLM's can be viewed as  approximate knowledge retrieval engines trained on Internet-scale data, and there has been several works that claim that the embedded knowledge in the network can solve planning tasks directly or through appropriate prompting~\cite{AGISparks23,wei2022emergentabilitieslargelanguage,LLM-Planner-23}. 
A less optimistic view is presented in~\cite{kambhampatiposition} which  reported that GPT-4 solves about 30\% of Blocksworlds puzzles which requires the LLM to generate a sequence of pick and place actions to reach a target configuration of  blocks from an initial disorganized state; Gemini Pro  had a success rate of only 0.5\%.
%\textcolor{blue}{Check the reference. These may be in different papers.} \textcolor{violet}{I am not sure where this is stated, did you mean Claude-3-Opus in \cite{stechly2024chainthoughtlessnessanalysiscot}  } 
A flurry of follow-on works have studied the capabilities of LLMs in Blocksworld and other common household planning benchmarks~\cite{LLM-Planner-23}.  Chain-of-thought (COT)  prompting, which supposedly ``unlocks the LLMs abilities to reason''~\cite{NEURIPS2022_COT}, has been applied to planning in~\cite{LLM-Planner-23} with some improvements. 
However,
in~\cite{stechly2024chainthoughtlessnessanalysiscot} 
the authors reaffirm the findings of~\cite{kambhampatiposition}  showing that meaningful performance improvements from COT prompts appear only when those prompts are hyper-specific to the problem,  and the improvements quickly deteriorate as
the number of blocks grow. See Section~\ref{sec:related-work}, for a discussion of other related work on path planning and LLMs.
\begin{figure}[h]
    \centering
    \includegraphics[width=0.24\linewidth]{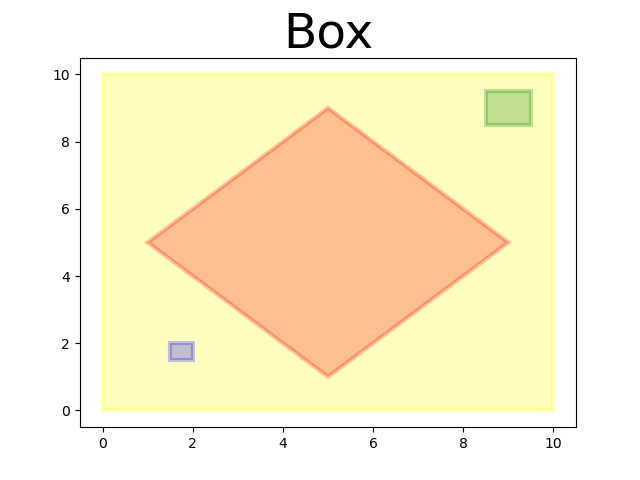}
    \includegraphics[width=0.24\linewidth]{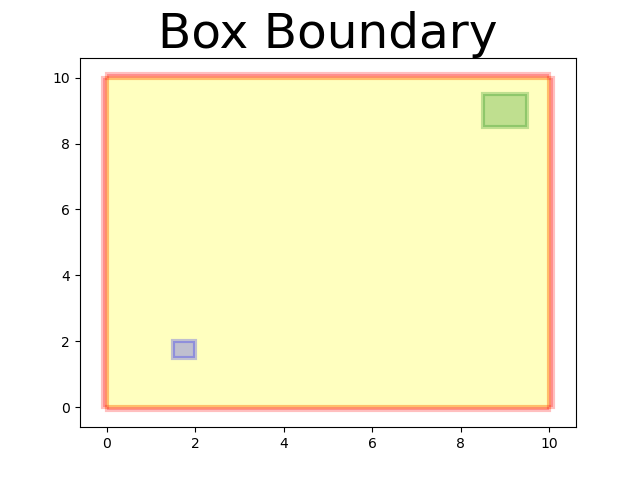}
    \includegraphics[width=0.24\linewidth]{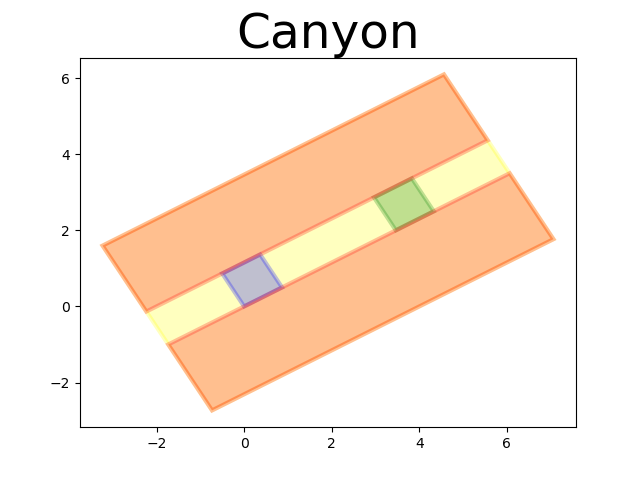}
    \includegraphics[width=0.24\linewidth]{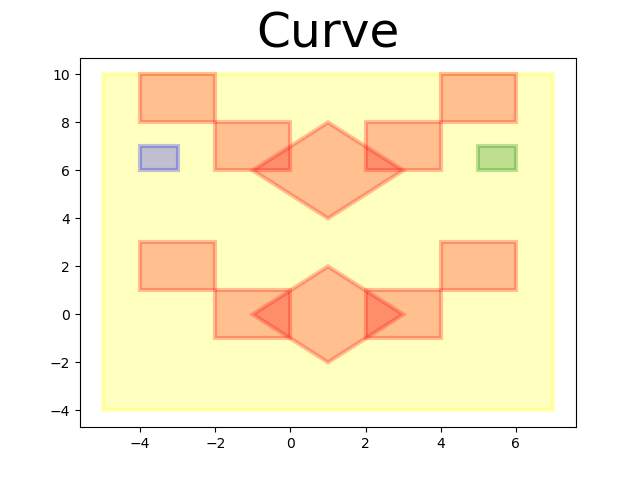} \\
    \includegraphics[width=0.24\linewidth]{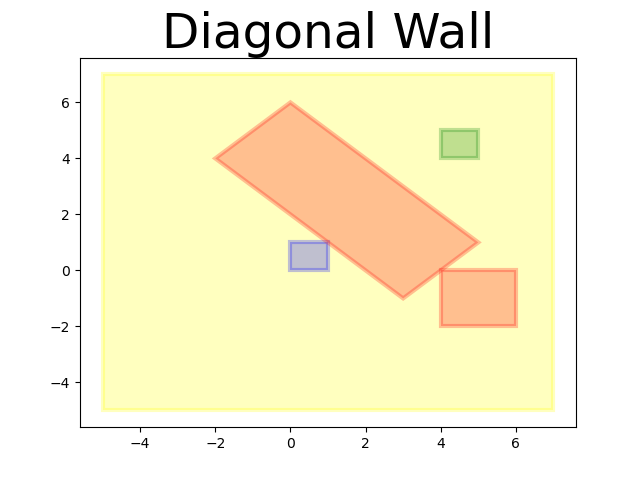}
    \includegraphics[width=0.24\linewidth]{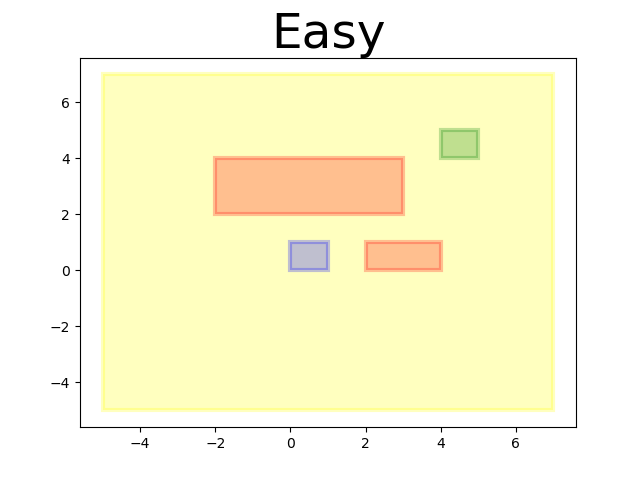}
    \includegraphics[width=0.24\linewidth]{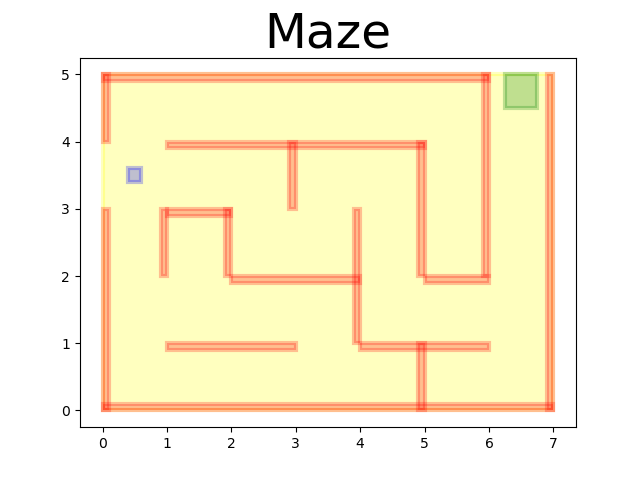}
    \includegraphics[width=0.24\linewidth]{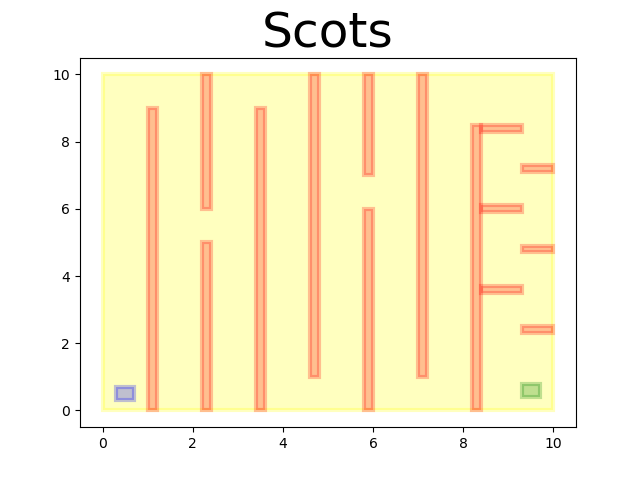}
    \includegraphics[width=0.24\linewidth]{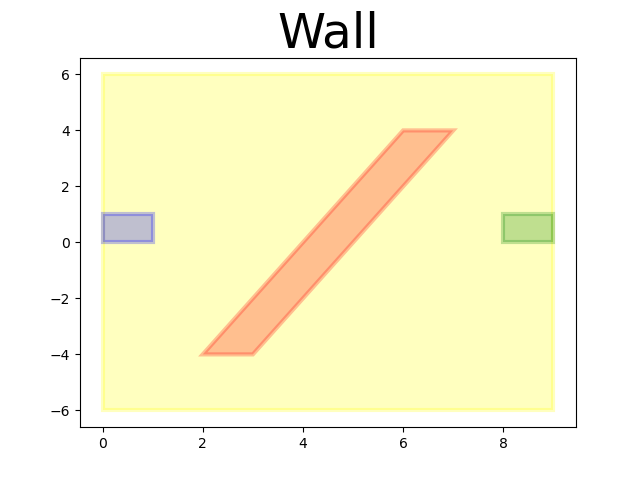}
    \includegraphics[width=0.24\linewidth]{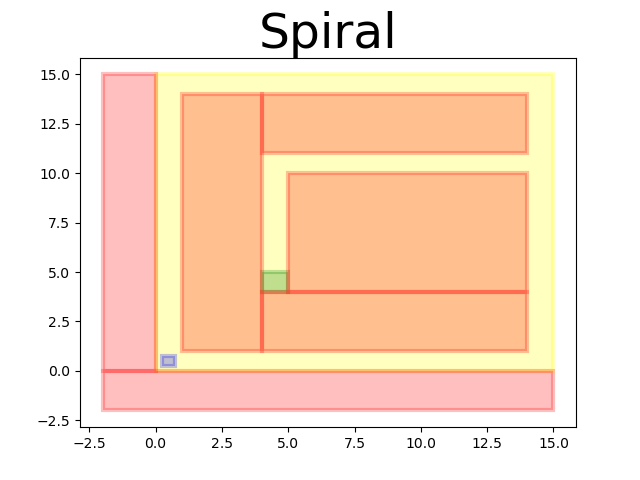}
    \caption{\small 2D path planning problems with initial set (blue), goal (green), and obstacles (red).}
    \label{fig:sample-maps}
\end{figure}

In this paper, we investigate the capabilities of LLMs in solving the classical {\em path planning} problem:  a robot or an agent has to find  a path from an initial position to a final goal, without colliding with obstacles or forbidden configurations. 
Deceptively simple to state path planning problems can be computationally hard,
and the  difficulty of problem instances usually grow with higher dimensions, with nonlinear agent dynamics, and with complex arrangement of obstacles.
We refer the interested readers to~\cite{Lavalle-planning06} for a discussion of lowerbounds and PSPACE-hardness of  planning problems.
There are many algorithms and software tools for solving this problem such as A$^*$~\cite{4082128}, RRT~\cite{kuffner2000rrt}, PRM~\cite{kavraki1996probabilistic}, RRG~\cite{karaman2009sampling}, just to name a handful.
 Path planning and the related problem of trajectory planning remain one of the most active area of robotics research~\cite{mendes2017real, herbert2017fastrack, Miller-FACTEST20}. In this paper, we study the performance of LLMs on a modest class of 2-dimensional path planning problems.

% 
% Hardness of problems, 
% Different strategies,
%\textcolor{orange}{
We selected 10 planning problems with different levels of difficulty from the literature (see Figure~\ref{fig:sample-maps}) and also use   100 randomly generated problems (Figure~\ref{fig:sample-maps-rand}). We study the performance of three LLMs---GPT-4o, Gemini Pro 1.5, Claude 3.5 Sonnet (see Section \ref{sec:implementation})---on this problem set with different kinds of closed-loop hinting strategies. LLMs cannot solve most of these problems without hints. 
We implemented a  framework in which the LLM's proposed solution path is analyzed by an satisfiability modulo theory (SMT) solver~\cite{Z3DeMoura} to generate four different  kinds of hints: collision hints say where (if at all) the path intersects with obstacles, free space hints point to possibly useful obstacle-free spaces, a prefix hint affirms a subpath that does not violate any constraints, and finally, an image hint provides a visualization of the puzzle. We experiment with different combinations of these hinting strategies and also with fine-tuning the LLMs~\cite{openai_gpt4_finetuning}.
%}
%However, this flexibility also comes with a downside: LLMs are prone to hallucination~\cite{ji2023towards}, which can lead to imprecise solutions—an issue that poses a significant challenge in tasks requiring accuracy, such as path planning.

%In our approach, we capitalize on the reactive nature of LLMs by integrating a feedback loop. After an initial attempt at solving a path planning task, an external solver is used to evaluate the proposed solution. Based on this evaluation, corrective hints are provided, allowing the LLM to reflect on its previous output and refine its path iteratively. 

\begin{figure}[h]
    \centering
    \includegraphics[width=0.32\linewidth]{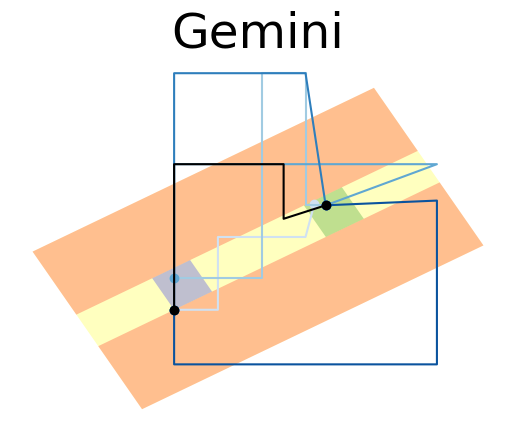}
    \includegraphics[width=0.32\linewidth]{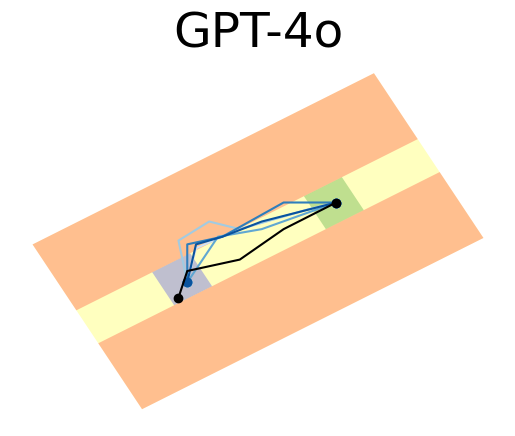}
        \includegraphics[width=0.32\linewidth]{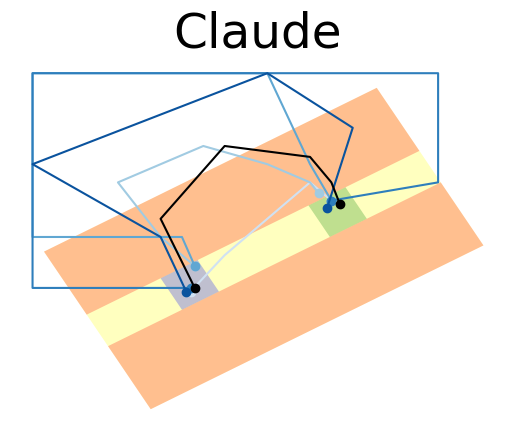}
    \caption{Examples of different LLM solutions using collision hints are shown. Gemini and Claude failed to solve the Canyon problem with collision hints after 20 feedback iterations (every 4th path shown). GPT-4o successfully solved the problem with collision hints in 6 iterations (all paths shown). In both cases, darker paths represent higher iteration numbers, with the final path depicted in black.}
    \label{fig:llm-bias-comparison}
\end{figure}

We found that solver-generated hints do improve  the LLM's ability to solve  path-planning problems, across the board. For example, with collision hints, Claude 3.5 Sonnet consistently achieves a success rate of  90\% for problems involving 1, 2, and 3 obstacles (Table \ref{table:collision_hints}), and Gemini finds a correct solution for the challenging Maze (Tables~\ref{table:collision_hints} and \ref{table:all_hints}). Overall, collision hints alone were adequate for LLMs to solve the moderately simple problems consistently. For the hardest problems in the set (which require ~25 segment paths and backtracking), the LLMs almost never managed to find a solution, with only a single instance of success, even with the maximum level of hint information. Notably, image hints did not have any marginal benefits in solving these problems.
% especially for simpler tasks, though even with guidance, models rarely succeeded in more difficult scenarios. Notably, collision-related hints were nearly as effective as adding more comprehensive ones, while visual hints did not aid performance, indicating limitations in processing visual data. 
We also found that fine-tuning models on path planning tasks significantly enhanced their ability to solve both the moderate and hard problem instances. GPT-4o improved its success rate for Box from 50\% to 100\% and for Diagonal Wall from 0\% to 30\%. We observed significant variations in the performance of the different LLMs. For example, Gemini and Claude struggled with tasks requiring diagonal movement, such as in the Canyon and the Curve problems, while GPT-4o demonstrated higher success rates in these  by generating  non-orthogonal paths (see Figure~\ref{fig:llm-bias-comparison}).
%
%
% Key findings:
% \textcolor{blue}{(1) WE confirm the findings of~\cite{} Wo hints none of the  LLMs cannot solve path planning beyond the simplest problems (). 
% %
% (2) With hints LLMs can solve the easy and the moderate problems most of the times and the harder problems rarely even with hints.
% %
% (3) More hints are better, but even with just collision hints the performance is as good as getting all hints. 
% (3.5) Images hints did not help, interestingly. 
% (4) Fine-tuning helps and improves the quality of the solutions.
% %
% (5) There is noticeable difference in the performance of the different LLMs, caveat.
% }

The main contribution of the paper is the comprehensive analysis of three of the current leading LLMs with respect to a new class of planning tasks,  that is central to robotics. The analysis is based on augmenting the LLMs with  closed-loop solver-generated guidance and fine-tuning. 
    %We also examine the effect of fine-tuning LLMs on solving path planning.
%\item 
A second contribution is the  opensource software framework for the evaluation of LLMs. The framework connects with different LLM  APIs, generates  closed-loop hints using SMT solvers, and as the LLM technology advances it will facilitate rapid experimentation and reproducibility. The framework also  comes with a suite of 10 handcrafted and 100 randomly generated 2D benchmark planning problems covering a range of difficulty levels. 
%\end{enumerate}

%% file: 03_Method.tex
\section{Problem and Method}
\label{sec:method}

Path planning is a classical problem  in robotics~\cite{Lavalle-planning06,Karaman2011}. 
A problem instance is specified by a initial position or configuration $\init$, a goal position $\goal$, and a finite collection of obstacles $\obs{1}, \ldots, \obs{N}$ or prohibited configurations.  
A sequence of positions (or configurations)  in the robots workspace is called a {\em path\/}. A path $w_1, \ldots, w_k$ is  {\em correct} if it starts at  $\init$, ends at $\goal,$ and for each $i<N$, the straight line joining $w_i$ and $w_{i+1}$ is disjoint from all obstacles $(\obs{j}'s)$. 
The length of the path $\sigma$ is the number of segments in the path or equivalently $k-1$.
Every set (e.g. $I$, $G$, and $O_1, ..., O_N$) is a convex polytope.
An algorithm {\em Alg} solves path planning if, given any problem instance it can find a correct path, provided one exists, and can decide that none exists if that is the case. 
The agent's workspace $\W$ is two or three-dimensional euclidean space for vehicles and can be higher-dimensional in general. 

In this paper, we empirically investigate whether the current generation of LLMs can solve path planning for two dimensional workspaces with quadrilateral obstacles (See Figure~\ref{fig:sample-maps} for some examples). Since every \emph{path} is described as a sequence of 2D coordinates in this setting, we refer to the positions $w_1, \ldots, w_k$ as \emph{waypoints}. The difficulty of a path planning problem  depends  factors such as the number of obstacles, the number of possible solutions, the robustness of solutions, etc. We use the path length as a proxy of difficulty of the problem as well as the the optimality of the found solution.

\begin{figure}
    \centering
    \includegraphics[width=\linewidth]{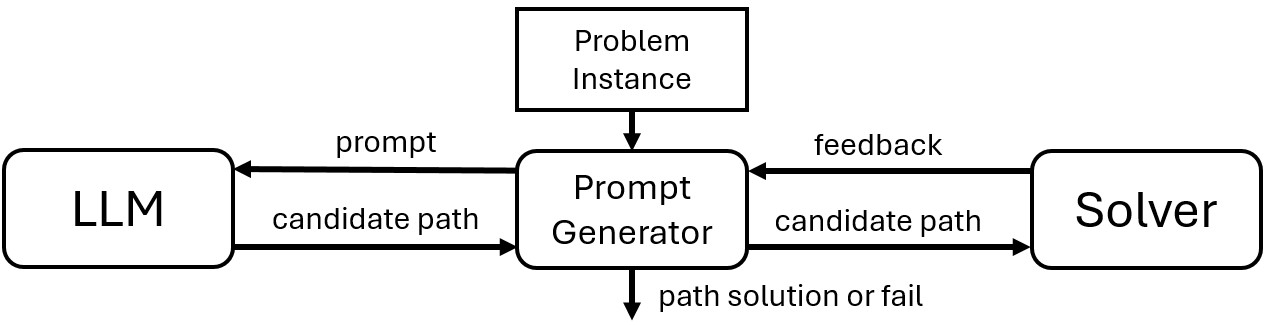}
    \caption{Flowchart of prompt generation, LLM, and solver feedback loop used in our analysis.}
    \label{fig:enter-label}
\end{figure}

\subsection{Closed Loop Prompting}
\label{sec:prompts}
Other researchers have noted that LLMs alone cannot reliably complete general planning and reasoning tasks~\cite{kambhampatiposition,guan2023leveraging}. This matches our experience with the path planning problem. If we explain the objectives of the path planning and prompt the LLMs only once to solve a problem instance, it almost never  generates a correct solution.  
%LLMs on their own are not exact and rather exhibit properties of approximate Knowledge Sources as approximate models of world/domain dynamics \cite{guan2023leveraging} and cannot generate executable paths on their own \cite{kambhampatiposition}.
%
With this observation, we  explore whether  LLMs  can solve planning problems with a sequence of prompts and with additional fine-tuning. 

An initial prompt describes the {\em generic\/} planning problem and introduces the elements specifying the problem such as the initial set, the goal set, and the obstacles. Each set is a quadrilateral specified by its vertices.
The initial prompt also contains instructions on the syntax of how the LLM should respond, that it should explain its thought process and one example problem with solution. 
% Earlier research suggests that prompting LLMs to outline their reasoning and generate intermediate steps, a method known as Chain of Thought, improves their performance~\cite{wei2022chain, NEURIPS2022_8bb0d291}. 
Providing LLMs with a few examples, known as few-shot prompting, has  proven to be effective~\cite{singh2023progprompt,cheng2022binding,ni2023lever}.

After the LLM generates an initial candidate solution, which is typically incorrect, the Prompt generator 
produces a sequence of prompts with additional hints to aid the LLM in solving the problem. These hints are calculated exactly by solving a set of linear constraints. The different kinds of hints are discussed in the next section. 
This prompting procedure continues for a number of iterations at the end of which, either a correct path is successfully found or we register a failure for this problem instance. 
%

% \textcolor{blue}{Is this describing fine-tuning? Just one correct example? } \textcolor{violet}{No, I included one example problem with solution with every prompt for "1-shot in-context" learning. Finetuning is trained on the whole model vs context given with each prompt here }
% In addition to this iterative prompting approach, we also utilized a 1-shot in-context learning method, where the LLM is provided with one example that includes a correct solution.
% %
% \textcolor{blue}{What does 1-shot in-context learning mean? These words need to be introduced / defined (if necessary) before we use them. What will happen if we did not provide this 1 example but gave the iterative hints?} 
% %
% The 1-shot approach allows the LLM to reference the structure and logic of a known correct solution, helping it to better understand the requirements of the task. This method has shown promising results in embodied agents \cite{singh2023progprompt}, question answering \cite{cheng2022binding}   and code generation \cite{ni2023lever},  where models demonstrated improved performance on various tasks by leveraging a single example.

\subsection{Hinting Strategies}
\label{sec:hints}
We  experimented extensively with the following hinting strategies and we report only a summary of the findings in  Section~\ref{sec:results}. We note that all hinting strategies are based on local checks that can be implemented correctly and efficiently using external constraint solvers.

\paragraph*{\textbf{Collision Hints}}
One type of useful  feedback an LLM can receive is a {\em local\/} explanation of how the proposed solution is incorrect. Specifically, given a proposed path $\sigma$ (generated by the LLM) the {\em collision hints } will include  the following types of information:
\begin{itemize} 
    \item whether $\sigma$  begins in the initial set $I$, 
    \item whether $\sigma$ ends in the goal set $G$, 
    \item which, if any, of the segments in $\sigma$ intersects  with obstacles.
\end{itemize}
Collision hint information is generated using an external constraint solver. Since \( I \), \( G \), and \( O_1, \dots, O_N \) are all convex polytopes, we can utilize an SMT solver like Z3~\cite{Z3DeMoura} to verify 
all of the above conditions. To see the details of how such constraints can be encoded as SMT problems, we refer the reader to~\cite{Miller-FACTEST20}.

\begin{figure}[h]
    \centering
    \includegraphics[width=0.4\linewidth]{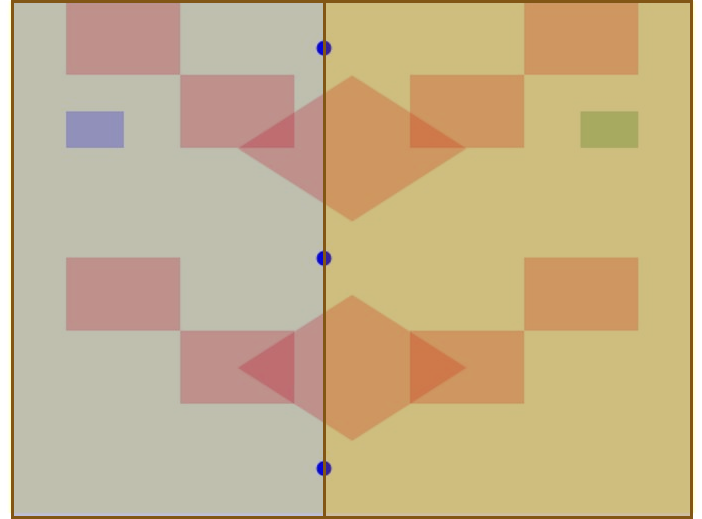}
    \hspace{0.05\linewidth}
    \includegraphics[width=0.4\linewidth]{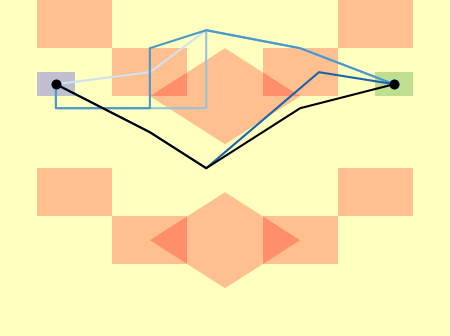} 

    \caption{On the left, the blue points are the free space hints showing possible points a correct path can go through and they are generated by vertically slicing the space. GPT-4o utilizes the free space hints on the right to find a solution.}
    \label{fig:curve_fsh}
\end{figure}

\paragraph*{\textbf{Free-Space Hints}}
We observed that often LLM generated paths  explore the shortest way to the goal $G$, and fail to explore alternative routes. To mitigate this problem  we provide it with a set of alternative waypoints, called {\em Free-Space Hints\/}. We split the workspace into multiple vertical slices and provide safe waypoints in each slide (see Figure \ref{fig:curve_fsh}). These free space hints define a tree of possibilities  that the LLM could  use to generate new paths.

\paragraph*{\textbf{Correct-Prefix Hints}}
To encourage the LLM to continue exploring correct prefixes of candidate paths, our third hinting strategy provides the longest prefix of an LLM-generated path $\sigma$ which starts in $I$ and does not violate the collision constraints. We observed that this helps steer the LLM toward more promising solutions without completely disregarding its previous answer.

\paragraph*{\textbf{Image Hints}}
With the  advances in multi-modal models, it is now possible to provide images as inputs to LLMs in addition to textual inputs.
Multi-modal models have been successfully applied in  robotics \cite{driess2023palmeembodiedmultimodallanguage} and image classification \cite{menon2022visualclassificationdescriptionlarge, DBLP:journals/corr/abs-2103-00020}.
Our fourth  strategy utilized this feature and provides an image rendering of the path planning problem (as in Figure~\ref{fig:sample-maps}) as an additional hint to the LLM. Picture hints involve providing the LLM with an image of the problem, alongside the last path it suggested. 
%This addition leverages the LLM's ability to process both visual and textual information, mimicking how humans intuitively solve 2D problems through visual inspection rather than relying solely on textual descriptions.

% . These advanced models combine textual and visual data, enabling more sophisticated interactions and understanding across different modalities, and have quickly found successful application in various fields such as embodied robots \cite{driess2023palmeembodiedmultimodallanguage} and image classification \cite{menon2022visualclassificationdescriptionlarge, DBLP:journals/corr/abs-2103-00020}.

\subsection{Fine Tuning}
\label{sec:finetuning}
Fine-tuning  is a technique to improve the performance of pre-trained LLMs with application-specific data~\cite{ding2023parameter}. This process involves  supervised traning on application-specific tailored dataset that is tailored to the desired application.  
%The primary advantage of fine-tuning is that it enables LLMs to achieve strong performance across a variety of benchmarks by leveraging the foundational knowledge acquired during the initial training phase. 
Fine-tuning has been shown to make LLMs more effective and accurate, as demonstrated in fields like finance ~\cite{yang2023fingptopensourcefinanciallarge} and medicine ~\cite{wu2024pmc}.
%~\cite{brown2020languagemodelsfewshotlearners, j2024finetuningllmenterprise}.
%
% Finetune examples: FinGPT (Finance) and PmC LLama (Medical) 
% Xiao-Yang Liu, Guoxuan Wang, Hongyang Yang, and Daochen Zha. Fingpt: Democratizing internet-scale data
% for financial large language models, 2023.
%  Chaoyi Wu, Weixiong Lin, Xiaoman Zhang, Ya Zhang, Yanfeng Wang, and Weidi Xie. Pmc-llama: Towards
% building open-source language models for medicine, 2023
%
For fine tuning LLMs for path planning, we construct a  dataset of problem-solution pairs by first randomly generating 200 planning problem instances following the same method outlined in Section~\ref{sec:problems} and generating correct solutions using the FACTEST tool~\cite{Miller-FACTEST20}. 
These fine tuning instances are distinct from the ones used for evaluations in~\ref{sec:results}.

%% file: 06_Results.tex
%%%%%%%%%%%%%%%%%%%%%%%%%%%%%%%%%%%%%%%%%%%%%%%%%%%%%%%%%%%%%%%%%%%%%%%%%%%%%%%%
\section{Experimental results}
\label{sec:experiments}
We will discuss the effectiveness of three different LLMs in solving path planning problems with different hinting strategies. Before presenting the results we discuss certain relevant details of the experiments.

\subsection{The Problems, metrics, and evaluation}
\label{sec:problems}
We evaluate the performance and impact of different hint strategies on two classes of path planning problems:

%\begin{enumerate}
%\item 
\paragraph*{Handcrafted Problems}
These are natural problems  obtained from existing literature on planning and control synthesis~\cite{Miller-FACTEST20,rungger2016scots}. They vary in the number of obstacles and the difficulty, i.e., the minimum solution length. All 10 handcrafted problems are shown in Figure~\ref{fig:sample-maps} and are listed on the top part of Tables, roughly in the order of increasing difficulty. 
LLMs paths are randomly generated, and therefore, for each handcrafted problem, we tested each hinting strategy by performing 10 separate evaluations. During these evaluations, the number of feedback iterations was capped at a maximum of 20, and we report the average success rates and path lengths. 

%were manually created with varying levels of difficulty to test different planning capabilities of the LLMs. Most of the problems were adapted from \cite{10.1007/978-3-030-53288-8_31}, while the "Scots" \textcolor{violet}{rename??} problem was inspired by \cite{rungger2016scots} .
    
\begin{figure}[h!]
    \centering
    \includegraphics[width=0.25\linewidth]{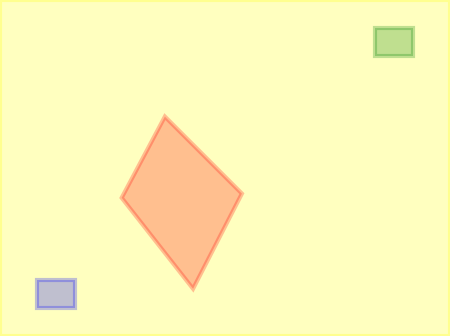}
    \hspace{0.04\linewidth} 
    \includegraphics[width=0.25\linewidth]{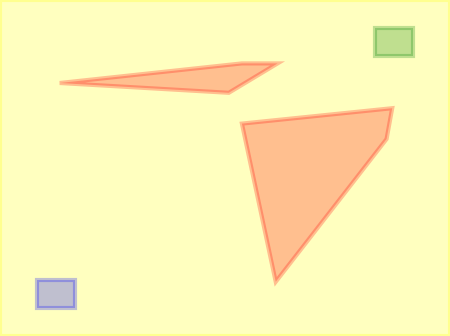}
    \hspace{0.04\linewidth} 
    \includegraphics[width=0.25\linewidth]{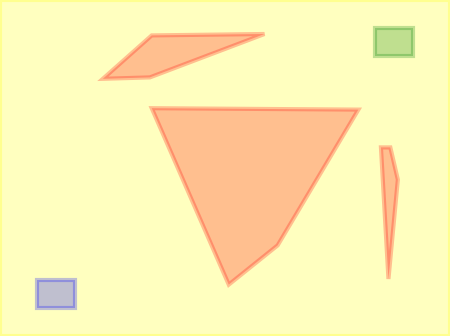} \\
    \vspace{0.04\linewidth} 
    \includegraphics[width=0.25\linewidth]{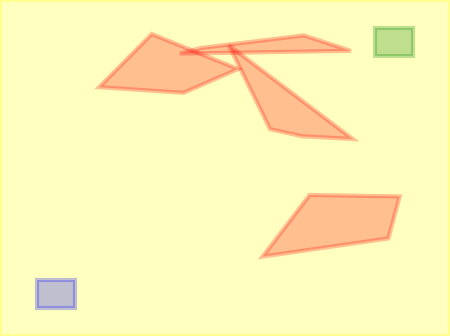} 
    \hspace{0.04\linewidth} 
    \includegraphics[width=0.25\linewidth]{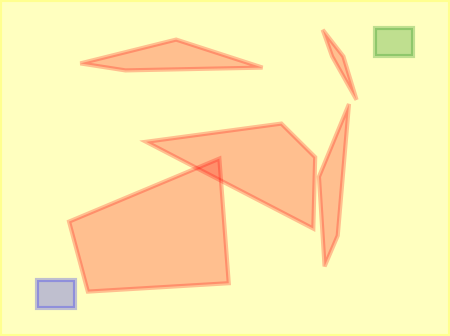}
    \caption{Examples of  randomly generated 2D path planning problems}
    \label{fig:sample-maps-rand}
\end{figure}

%\item 
\paragraph*{Random  Problems}
We have generated a large collection of random problems for this work,
For a given number of obstacles $k$, the workspace is divided into a grid of $n>K$ tiles. A hyperparameter controls the extent to which these tiles can overlap. Out of the $n$ tiles, $k$  are randomly selected, and four random points within the tile are generated, which form a convex obstacle. This approach can generate planning problems with different difficulty levels by varying the number, the locations,  and the overlaps of the obstacles in the workspace. Several examples from this class of problems are shown in Figure~\ref{fig:sample-maps-rand} and are listed on the lower part of Tables. 

We evaluate  randomly generated problems with $n=1, \ldots, 5$ obstacles. For each obstacle count, we evaluated the different hinting strategies on 20 distinct randomly generated problem instances. This means, for example, that we tested on 20 different problems with 1 obstacle, 20  problems with 2 obstacles, and so on. For each problem, the maximum number of feedback iterations was limited to 5,
which differs from the 20-iteration cap used in handcrafted problems. This is because the random problems tend to be simpler and our aim here was to study a range of problems rather than multiple attempts on the  same problem.

%This distinction in evaluation methods reflects the differing nature of the problem types. Handcrafted problems, being specifically designed, allowed for repeated testing on the same problem and better traceability, whereas randomly generated problems necessitated testing on multiple instances to ensure robust evaluation across varying scenarios.

%For the tables demonstrating the results, we roughly arranged the handcrafted problems according to our subjective assessment of their difficulty.

\paragraph*{Performance Metrics}
\label{sec:metrics}
For each experiment, we report three key metrics: Success rate (S\%), the average number of iterations required to find a correct path (N), and the average length of the correct path (PL).
The success rate  measures the proportion of problems in which the hint strategy successfully guided the LLM to find a correct solution. The average number of iterations captures the efficiency of the hinting process. This metric is also a proxy for the running time. Finally, the average path length reflects the complexity of the solution. Together these metrics collectively provide a different perspectives on the effectiveness of the different hinting  strategies.

\subsection{Implementation and the LLMs}
\label{sec:implementation}

Our iterative hint-generation framework is implemented in Python 3.10 using  Z3 as the solver~\cite{Z3DeMoura}. The code will be made publicly available.

\paragraph*{LLMs} We tested a variety of language models, including locally-hosted options like Llama 3 \cite{dubey2024llama3herdmodels} and Mistral NeMo \cite{Mistral2024NeMo}. However, in what follows we present the detailed results from proprietary models owing to their superior performance. The models used in our experiments were Gemini-Pro-1.5-001 \cite{geminiteam2024gemini15unlockingmultimodal}, ChatGPT-4o-2024-08-06  \cite{openai2024gpt4technicalreport}, and Claude-3.5-Sonnet-20240620 \cite{Claude3Addendum2024}. In the following text, they will be referred to as Gemini, GPT-4o, and Claude (Claude 3.5 Sonnet). These models were accessed via API, utilizing the default parameters provided by the respective platforms. The temperature parameter was left at its default value, rather than being set to 0 as commonly done in other studies, to encourage more diverse responses~\cite{peeperkorn2024temperaturecreativityparameterlarge}. As mentioned, we perform multiple experiments for each setup to report average performance metrics. % higher temperature led to better performance in LLM-Assist 

\paragraph*{Local Computations} The experiments and prompt generation were executed on a local machine running Ubuntu 22.04 LTS. The machine was equipped with an NVIDIA GeForce GTX 1660 SUPER GPU (6 GB VRAM), an AMD Ryzen 5 3600 6-core processor, and 16 GB of RAM. Note that this machine was used for managing the experiments and handling auxiliary tasks, the LLM inferences ran on the servers of the respective API providers.

% The running time for each experiment was measured from the initiation of the first prompt to the final file write. This time measurement is influenced by several factors, including the latency of the APIs used to access the models and the write speed of the machine, as all outputs were logged throughout the experiment. While this metric is dependent on these variables, it nonetheless provides a general indication of the time required to conduct the experiments.

\subsection{Main Results}
\label{sec:results}

We  evaluated the effectiveness of the different hinting strategies listed in Section~\ref{sec:hints} and their combinations. 
Without any hints, we found that all the LLMs struggled to solve path planning problems beyond the simplest cases. This is consistent  with the findings of~\cite{kambhampatiposition,guan2023leveraging}.
We proceed to report the results from three different levels of hints that cover the other combinations.

\subsubsection{LLMs with collision hints can solve moderate but not the hard planning problems}
\label{sec:results_ch}
Collision hints provide  local feedback, which makes them a baseline for the other hinting strategies. 
The introduction of collision hints improved the performance of all the LLMs, particularly for easy and moderate problems (see Table~\ref{table:collision_hints}). However, LLMs with collision hints rarely succeeded with the more difficult problems like Maze and Scots. Claude  solved Scots once in 10 attempts.

There is substantial differences across the LLMs.  Gemini and Claude tend to generate axis-aligned paths, while GPT-4o frequently produces paths with diagonal segments (see Figure~\ref{fig:llm-bias-comparison}). 
GPT-4o consistently outperformed Gemini and Claude on the Canyon problem but was completely unable to solve Diagonal Wall. In addition, Gemini was the only LLM that never solved Box Boundary on the first try. These failures were  primarily because the proposed solution did not reach the  goal $G$. Overall, Claude demonstrated the strongest performance across the board.

For the random problems, no big surprises: with more obstacles, the problems became harder for the LLMs. Claude performed well, with over 90\% success rate on problems with up to three obstacles. However, its performance dropped to only 25\% success rate when faced with four or more.

\begin{table}[h]
\centering
    %\hspace*{-0.25cm}
    \setlength{\tabcolsep}{5pt}
    \begin{tabular}{lccccccccc}
    \toprule
    & \multicolumn{3}{c}{Gemini Pro 1.5} & \multicolumn{3}{c}{GPT-4o} & \multicolumn{3}{c}{Claude 3.5 Sonnet} \\ 
    \cmidrule(lr){2-4} \cmidrule(lr){5-7} \cmidrule(lr){8-10}              
    \textbf{Problem}    & \textbf{S\%} & \textbf{N} & \textbf{PL} 
                        & \textbf{S\%} & \textbf{N} & \textbf{PL} 
                        & \textbf{S\%} & \textbf{N} & \textbf{PL} \\ 
    \midrule
    \textbf{Box B.}          & \textbf{100} & 1.3  & 3.7   & \textbf{100} & 1.0  & 3.0  & \textbf{100} & 1.0  & 5.3  \\ 
    \textbf{Easy}            & \textbf{100} & 3.3  & 4.5   & \textbf{100} & 2.6  & 5.2  & \textbf{100}   & 3.2  & 4.9 \\ 
    \textbf{Wall}            & \textbf{100} & 1.6  & 4.2  & \textbf{100} & 1.6  & 4.5  & \textbf{100}  & 1.5  & 4.8  \\ 
    \textbf{Box}             & 90  & 10.9  & 4.4  & 50  & 8.6 & 5.2  & \textbf{100}  & 2.2 & 5.2  \\ 

    \textbf{Canyon}          & 0  & -  & -        & \textbf{100} & 2.2  & 4.7  & 40  & 10.0  & 6.5  \\ 
    \textbf{D. Wall}         & \textbf{70}  & 8.9  & 5.0   & 0   & - & - & 40   & 10.8  & 5.5 \\ 
    \textbf{Curve}           & 40  & 8.0 & 6.5    & \textbf{70} & 9.1 & 5.4  & 60  & 9.0 & 7.8  \\ 
    \textbf{Spiral}          & 60  & 14.2 & 6.2  & 0   & -  & -  & \textbf{80}   & 14.4  & 9.4 \\ 

    \textbf{Maze}           & 0   & -     & -    & 0   & - & -  & 0   & -  & - \\ 
    \textbf{Scots}           & 0  & - & -  & 0   & -  & -  & \textbf{10}   & 11.0  & 30.0 \\

    \midrule
    \textbf{1 Obs}           & \textbf{95}  & 1.7  & 4.1  & 90  & 1.7  & 4.9  & \textbf{95}  & 1.5  & 6.2  \\ 
    \textbf{2 Obs}           & 95  & 2.5  & 4.9  & 50  & 2.3  & 5.7  & \textbf{100}  & 2.8  &  7.3  \\ 
    \textbf{3 Obs}           & 75  & 2.9  & 4.9  & 15  & 4.0  & 6.3  & \textbf{90}  & 3.4  & 8.1  \\ 
    \textbf{4 Obs}           & \textbf{45}  & 3.7  & 4.9  & 5   & 1.0 & 5.0  & 25 & 2.4  & 9.4 \\ 
    \textbf{5 Obs}           & \textbf{30}  & 3.7  & 4.8  & 10  & 3.0  & 7.0  & 15  & 4.0  & 9.0  \\ 
    \bottomrule
    \end{tabular}
\caption{LLM performance with collision hints. The average success rate (S\%), number of hints (N), and length of the found path (PL) needed to find a correct solutions.}
\label{table:collision_hints}
\end{table}

\subsubsection{Collision, free space, and correct prefix hints improve performance for handcrafted problems, but shows mixed results for Spiral and random problems}
\label{sec:results-all-text-hints}

Combining collision hints with free space and correct prefix hints gives the LLMs the most extensive amount of information  that our text-based hinting strategies allow. The results are shown in Table~\ref{table:all_hints}. 
With  more extensive guidance, the LLMs  demonstrated an improved performance across nearly all handcrafted problems, with the exception for Spiral. Performance was equal or better than before for the majority of tasks.  Gemini, which  could not solve Canyon with only collision hints, improved to a 30\% success rate; Claude's success rate increased from 40\% to 90\%;  GPT-4o  consistently solved Curve.
For hard problems like Maze, the free space with correct prefix hints were critical. Correct prefix hints enabled the LLMs to refine previous candidate solutions  rather than discarding them  entirely. Without prefix hints, the iterative feedback can fall into random guessing.

Interestingly, the additional hints resulted in a {\em worse} performance for Spiral. Gemini's success rate dropped from 60\% to 30\%, and Claude's performance declined even more sharply, from 80\% to 20\%.
In the case of the Scots problem, Claude now encounters difficulties to solve the problem. It consistently finds a path through the suggested free space hint, but when compared to using collision hints only, the model now moves more directly, passing through obstacles to reach the free space hint, rather than taking the longer, but obstacle-free path.

The trend is unclear for the random problems. While Gemini and Claude achieve perfect success rates on one-obstacle problems, their success rate drop by at 35\% and 25\% for problems with two obstacles, respectively. There is no significant change in the success rate for GPT-4o.

Generally, the number of feedback iterations needed to find a correct solution is higher with collision, free space, and prefix hints for Gemini compared to only using collision hints. For the Wall problem, Gemini needs more than four times and for Curve twice as many iterations on average. On the other hand, GPT-4o not only improved in solving Curve with a 100\% success rate, it also needed half as many iterations. Claude demonstrated comparable improvements on the Canyon problem.

The mean length of correct paths increased in comparison to only using collision hints. This can be attributed to the inclusion of free space hints, which requires the path to incorporate at least one such point.
Overall, while offering more guidance generally led to higher success rates, it did not consistently improve optimality (Table~\ref{table:collision_hints}). 

\begin{table}[h!]
    %\hspace*{-0.3cm}
    \setlength{\tabcolsep}{5pt}
    \begin{tabular}{lccccccccc}
    \toprule
     & \multicolumn{3}{c}{Gemini Pro 1.5} & \multicolumn{3}{c}{GPT-4o} & \multicolumn{3}{c}{Claude 
     3.5 Sonnet} \\ 
    \cmidrule(lr){2-4} \cmidrule(lr){5-7} \cmidrule(lr){8-10}              
    \textbf{Problem}    & \textbf{S\%} & \textbf{N} & \textbf{PL} 
                        & \textbf{S\%} & \textbf{N} & \textbf{PL} 
                        & \textbf{S\%} & \textbf{N} & \textbf{PL} \\ 
    \midrule
    
    \textbf{Box B.}    & \textbf{100} & 1.3  & 3.7  & \textbf{100} & 1.0  & 3.9  & \textbf{100} & 1.0  & 4.9  \\ 
    \textbf{Easy}            & \textbf{100} & 3.9  & 6.6  & \textbf{100} & 6.4  & 5.6  & \textbf{100} & 2.2 & 6.2  \\ 
    \textbf{Wall}            & \textbf{100} & 8.0  & 6.5  & \textbf{100} & 5.4  & 6.9  & \textbf{100}   & 3.8 & 6.8  \\ 
    \textbf{Box}             & 90  & 8.8  & 6.3  & 50  & 17.3 & 8.0  & \textbf{100}  & 4.0 & 7.6 \\ 
    \textbf{Canyon}          & 30  & 15.9 & 5.3  & \textbf{100} & 2.6  & 4.6  & 90  & 2.4  & 7.1 \\ 
    \textbf{D. Wall}         & \textbf{90}  & 8.9  & 7.7  & 0   & -  & -     & 30   & 9.3     & 6.7     \\ 
    \textbf{Curve}           & 30  & 17.9 & 11.0 & \textbf{100} & 4.3  & 5.8  & 70  & 10.0 & 7.1 \\ 
    \textbf{Spiral}          & \textbf{30}  & 17.1 & 9.0  & 0   & - & -     & 20   & 14.0     & 14.0     \\ 
    \textbf{Maze}            & \textbf{10}  & 20.1 & 13.0 & 0   & - & -     & 0   & -     & -     \\ 
    \textbf{Scots}           & 0   & -     & -     & 0  & - & - & 0   & -     & -     \\

    \midrule
    \textbf{1 Obs}      & \textbf{100} & 1.9  & 4.8  & 70  & 1.8  & 5.1  & \textbf{100}  & 1.7  & 6.4  \\ 
    \textbf{2 Obs}     &  60  & 2.8  & 6.3  & 50  & 2.6  & 5.5  & \textbf{75}  & 2.4  & 8.0  \\ 
    \textbf{3 Obs}     & \textbf{65}  & 4.1  & 6.7  & 20  & 2.2  & 6.2  & 60  & 2.6  & 9.4  \\ 
    \textbf{4 Obs}     & \textbf{20}  & 5.0  & 7.5  & 10  & 5.5  & 9.0  & \textbf{20}  & 3.8  & 11.5  \\ 
    \textbf{5 Obs}     & \textbf{35}  & 3.8  & 6.8  & 15  & 3.0  & 7.0  & 15  & 2.33  & 8.0  \\ 
    \bottomrule
    \end{tabular}
\caption{LLM Performance with collision, free space and prefix hints.}

\label{table:all_hints}

\end{table}

\subsubsection{Image hints did not help LLMs solve planning problems}
\label{sec:results-image-hints}
Two-dimensional planning problems and their solutions can be visualized clearly. 
However, image hints,  did not appear to enhance the LLM's path-planning performance. 
The results of using all the textual hints of Section~\ref{sec:results-all-text-hints} together with image hints are shown in  Table~\ref{table:vision_results}.
This outcome is surprising, given that LLMs have demonstrated strong image comprehension in other contexts, including the ability to read text within images~\cite{liu2024ocrbenchhiddenmysteryocr} and understand visual context \cite{driess2023palmeembodiedmultimodallanguage, menon2022visualclassificationdescriptionlarge, DBLP:journals/corr/abs-2103-00020}.
In some cases, the LLM's performance even degraded with images.

\begin{table}[h!]
 %\hspace*{-0.25cm}
\setlength{\tabcolsep}{5pt}
\centering
\begin{tabular}{lccccccccc}
\toprule
 & \multicolumn{3}{c}{Gemini 1.5} & \multicolumn{3}{c}{GPT-4o} & \multicolumn{3}{c}{Claude 3.5 Sonnet} \\ 
\cmidrule(lr){2-4} \cmidrule(lr){5-7}  \cmidrule(lr){8-10}   
\textbf{Problem}  & \textbf{S\%} & \textbf{N} & \textbf{PL} 
                  & \textbf{S\%} & \textbf{N} & \textbf{PL} 
                  & \textbf{S\%} & \textbf{N} & \textbf{PL} \\
\midrule
\textbf{Box B}           & \textbf{100} & 1.0  & 3.0  & \textbf{100} & 1.0  & 3.4 & \textbf{100} & 1.0 & 5.0 \\ 
\textbf{Easy}            & \textbf{100} & 4.8  & 4.8  & 90 & 7.3  & 5.7 & \textbf{100} & 2.3 & 5.2\\ 
\textbf{Wall}            & 70 & 5.4  & 4.7  & \textbf{100} & 6.4  & 6.1 & \textbf{100} & 2.9 & 6.5 \\ 
\textbf{Box}             & \textbf{100} & 3.1  & 4.4  & 40 & 12.0 & 5.5 & \textbf{100} & 3.0 & 7.4\\ 
\textbf{Canyon}          & 50  & 8.8  & 4.6  & \textbf{100} & 2.0  & 4.4 & \textbf{100} & 3.3 & 6.3\\
\textbf{D. Wall}         & \textbf{40}  & 13.8 & 5.8  & 0  & - & -  & 30 & 9.7 & 7.3\\ 
\textbf{Curve}           & 60 & 14.5 & 6.3  & \textbf{90} & 4.6  & 4.9 & 40 & 5.0 & 8.8\\ 
\textbf{Spiral}          & \textbf{20}  & 12.0 & 6.5  & 0  & - & - & \textbf{20} & 14.5 & 12.0 \\ 
\textbf{Maze}            & \textbf{10}  & 10.0 & 11.0 & 0  & - & - & 0 & - & - \\ 
\textbf{Scots}           & 0   & -    & -    & 0  & - & - & 0 & - & -\\ 

\midrule
\textbf{1 Obs}      & \textbf{95}  & 1.6  & 3.4  & 75 & 2.0  & 4.4 & \textbf{95}  & 1.8  & 6.2  \\  
\textbf{2 Obs}     & \textbf{90}  & 2.6  & 3.8  & 40 & 2.6  & 5.0 & 80  & 2.6  & 7.8  \\  
\textbf{3 Obs}     & \textbf{55}  & 3.1  & 3.6  & 25 & 2.8  & 5.8 & 45  & 2.4  & 8.2  \\ 
\textbf{4 Obs}     & \textbf{50}  & 2.4  & 3.7  & 10  & 2.0    & 5.0   & 25   & 2.6  & 9.6  \\ 
\textbf{5 Obs}     & \textbf{35}  & 2.4  & 4.1  & 20 & 3.3  & 6.5 & 10  & 3.5  & 9.0  \\
\bottomrule
\end{tabular}
\caption{LLM Performance with collision, free space, prefix and image hints.}
\label{table:vision_results}
\end{table}

\subsubsection{Fine tuning improves optimality of  solutions}
\label{sec:results-fine-tuning}
We used the fine-tuning capabilities of GPT-4o for this experiment~\cite{openai_gpt4_finetuning, openai2024gpt4technicalreport}; other models did not offer fine-tuning functionality at the time of writing. Fine-tuning  (see Section~\ref{sec:finetuning}) led to a noticeable improvement in the length of the paths found (See Table~\ref{table:fine_tuning}).
GPT-4o is able to solve Diagonal Wall more often  and its performance on the random problems improved significantly. The paths that the LLM generated are generally shorter. 

The dataset used for fine-tuning did not include prompts for explaining solutions and relied solely on solution arrays. We also experimented with fine-tuning the models on a synthetic dataset, where explanations for correct solutions were generated by another LLM. However, this approach proved less effective than using the dataset that contained only the solution arrays, indicating that direct task-specific training data may be more beneficial for path-planning  than generated (and possibly wrong) explanations.

\begin{table}[h]
\centering
\begin{tabular}{lcccc}
\toprule
\textbf{Environment} & \textbf{S\%} & \textbf{N}  & \textbf{PL} \\ 
\midrule
\textbf{Box Boundary} & 100 & 1.5   & 3.1  \\
\textbf{Easy}         & 90  & 4.9   & 4.1  \\ 
\textbf{Wall}         & 100 & 7.3  & 3.9  \\ 
\textbf{Box}          & 100 & 2.9   & 4.0  \\ 
\textbf{Canyon}       & 100 & 4.6   & 3.3\\ 
\textbf{Diagonal Wall}& 30  & 10.7   & 4.0  \\ 
\textbf{Curve}        & 100 & 5.1    & 3.9  \\ 
\textbf{Spiral}       & 0  & -   & - \\ 
\textbf{Maze}         & 0   & -   & - \\ 
\textbf{Scots}        & 0  & -   & -  \\ 
\midrule
\textbf{1 Obs}   & 90  & 1.6    & 3.4  \\ 
\textbf{2 Obs}  & 90  & 2.6    & 3.8  \\ 
\textbf{3 Obs}  & 55  & 3.1    & 3.6  \\ 
\textbf{4 Obs}  & 50  & 2.4  & 3.7  \\ 
\textbf{5 Obs}  & 35  & 2.4   & 4.1  \\ 

\bottomrule
\end{tabular}
\caption{Fine-tuned GPT-4o performance using collision, free space and prefix hints}
\label{table:fine_tuning}
\end{table}

\subsubsection{Gemini and Claude tend to generate orthogonal paths}
\label{sec:results-llm-differences}
In our experiments, we observed differences in the path-planning tendencies and performance of the LLMs. Both Gemini and, to some extent, Claude demonstrated a bias towards generating orthogonal movements (Figure \ref{fig:llm-bias-comparison}).
For instance, the Canyon problem can be solved with two waypoints connected by a diagonal segment. Gemini and Claude often encountered difficulties with this problem  (Figure~\ref{fig:llm-bias-comparison}). Similar outcomes were observed in the Curve problem, where orthogonal paths presented more challenges.

By the same reasoning, Gemini and Claude have  better performance on problems like Diagonal Wall and Box. 
%It was observed that LLMs were more successful in avoiding obstacles by proposing paths that have fewer waypoints and move orthogonally around obstacles, than having more dense waypoints that move more freely through space. 
In the Maze and the Scots problem, axis-aligned paths also proved advantageous for avoiding obstacles (Figure \ref{fig:maze-scots-sol}). GPT-4o, on the other hand, frequently failed to recognize the obstacles, often moving directly through the middle.

%% file: 07_Conclusions.tex
\begin{figure}[h!]
    \centering
    \includegraphics[width=0.45\linewidth]{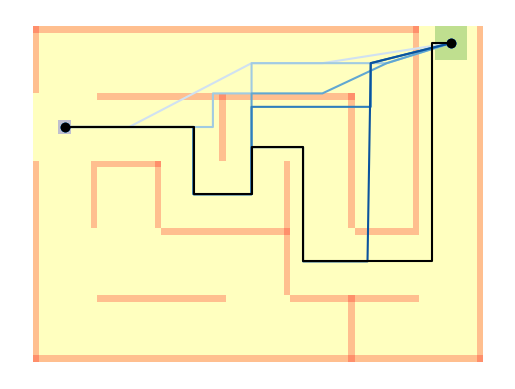}
    \includegraphics[width=0.45\linewidth]{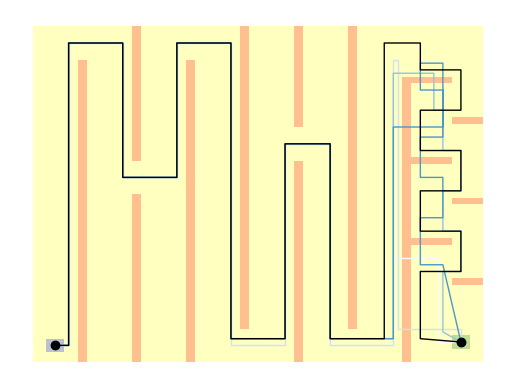}
    \caption{Examples of LLM evaluations on challenging problems are shown. On the left, Gemini found a correct solution for the Maze problem in 20 iterations (every 4th path shown). On the right, Claude successfully solved the Scots problem in 11 iterations (every 3rd path shown).% The coloring scheme is consistent with that in Fig. \ref{fig:llm-bias-comparison}.
    }
    \label{fig:maze-scots-sol}
\end{figure}

\section{Related Work}
\label{sec:related-work}
% There already exists a rich body of work surrounding classical planners for path planning. For example, 
% such as A$^*$~\cite{4082128}, RRT~\cite{kuffner2000rrt}, PRM~\cite{kavraki1996probabilistic}, RRG~\cite{karaman2009sampling}, etc. However, they are rule-based and do not use any kind of generative methods.
An alternative approach for integrating LLMs in the overall planning stack is to use them as an interface between human users specifying the problem and the (classical) algorithmic  planners. In this, the LLMs function as a translator from natural language to the formal or computer-readable specification of the task or motion planning problem and this has been explored in~\cite{liu2023llm+} and~\cite{10611163}.

A very different  approach for combining LLMs with path planners has been explored in~\cite{lin2024drplanner,meng2024llmalargelanguagemodel}. Here the idea is to correct or improve the performance of classical planners, with LLMs. In~\cite{lin2024drplanner} the LLM is used to diagnoses problems in a generated plan  from detailed information about the scenario, type of algorithm used, objective functions, etc.
In~\cite{meng2024llmalargelanguagemodel} a method is proposed for leveraging LLMs to guide search the classical A* algorithm.

Finally, we mention that there has been research  on closed-loop prompting with LLMs for travel planning~\cite{gundawar2024robust}, autonomous driving~\cite{fu2024limsim++, Shao_2024_CVPR}, and task planning~\cite{10610065}, which are less directly related to our work.

\section{Limitations and Future Work}
\label{sec:limitations}
% This spotlights drawbacks of chain of thought, especially the sharp
% tradeoff between possible performance gains and the amount of human labor
% necessary to generate examples with correct reasoning traces.

For planned paths to be useful, the should be path dynamically feasible. A correct path with sharp turns around an obstacle may not be safe for a vehicle with a large turning radius. Such dynamic or nonholonomic constraints are currently not included in our framework, although there are many algorithms for solving that  problem. We did not investigate these dynamic constrains here because the LLMs seem to struggle with hard path planning problems even without dynamic constraints. As the LLMs improve, in the future it will be important to incorporate dynamic constraints  and the constraints arising from agent's geometry into the planning task.

% planning is for designing paths and synthesizing controllers for non-linear vehicles. To generate realistic solutions, it is crucial to account for vehicle dynamics. Currently, our framework generates a series of waypoints connected linearly, which may be impossible for vehicles to follow, especially if the path includes sharp 90-degree turns. Unfortunately, as discussed in Section \ref{sec:results-llm-differences}, Gemini and Claude, in particular, tend to produce these kinds of linear paths, which can be challenging for vehicles to navigate effectively.

%Secondly, Another limitation of our framework is that its path-planning capabilities are currently inferior to those of traditional path planners, which generally achieve higher success rates and more efficient runtime. However, as an exploratory study, we aim to report on the current state of LLM path-planning capabilities and their potential. With advancements in reasoning capabilities, LLMs could see improvements in this framework and potentially narrow the gap with traditional planners in the future.

%Lastly, due to the nondeterministic nature of LLMs, establishing concrete constraints or guarantees regarding runtime or solution quality is challenging. This variability can complicate the assurance of reliability and consistency in the generated action plans. The inherent randomness may affect the ability to meet specific constraints or achieve desired outcomes, particularly in scenarios where dependable results are crucial. 

Additional prompting and fine-tuning strategies that are effective in other contexts should be explored for path planning as well. For example, presenting sequences with smaller reasoning steps, example solutions, identifying the symmetries and invariances that exist in the problem domain, building-in linear constraint checks, etc. Our software framework can indeed be extended to support several of these hinting strategies.

Finally, we conjecture that  attaching the planning problems to real world data that is supposedly embedded in the LLM will improve their performance. To test this, we will have to present planning problems in the real world with names of places, street, zones (instead of abstract  obstacles and free-space). To this end, it will be interesting to connect the planning problems with a geographical data and realistic maps.

\section{Conclusion}
\label{sec:conclusion}
In this work, we propose an approach to assist LLMs in complex planning tasks by incorporating solver feedback in a closed-loop system and evaluating their performance with different types of hints. We also introduce a benchmark of 10 handcrafted and 100 randomly generated problems. 
Our experiments show that even minimal collision hints significantly improve performance on moderately difficult problems, with LLMs consistently finding correct solutions once provided with the hints. However, LLMs still struggle with the most challenging tasks, even when provided with the full range of hints generated by our framework. Notably, image-based hints did not enhance performance, highlighting limitations in processing visual data. Fine-tuning, however, led to more consistent and efficient solutions. Finally, Gemini and Claude tend to generate orthogonal movements, in contrast to GPT-4o.
Future work will focus on incorporating dynamic constraints and agent geometry into path planning tasks, as well as exploring prompting strategies like smaller reasoning steps and real-world data integration to improve LLM performance.